\documentclass{article}

\usepackage{arxiv}

\usepackage[utf8]{inputenc}
\usepackage[T1]{fontenc}
\usepackage{hyperref}
\usepackage{url}
\usepackage{booktabs}
\usepackage{amsfonts}
\usepackage{amsmath}
\usepackage{nicefrac}
\usepackage{microtype}
\usepackage{graphicx}
\usepackage[numbers,sort&compress]{natbib}
\usepackage{doi}

\renewcommand{\headeright}{}
\renewcommand{\undertitle}{}

\title{Benchmarking Logistic Regression, SVM, Naive Bayes, and IndoBERT Fine-Tuning for Sentiment Analysis on Indonesian Product Reviews}

\author{
  Nabila Zakiyah Zahra \\
  Department of Data Science\\
  Institut Teknologi Sumatera (ITERA)\\
  \texttt{122450139@student.itera.ac.id} \\
  \And
  Salwa Farhanatussaidah \\
  Department of Data Science\\
  Institut Teknologi Sumatera (ITERA)\\
  \texttt{122450055@student.itera.ac.id} \\
  \AND
  Nasywa Nur Afifah \\
  Department of Data Science\\
  Institut Teknologi Sumatera (ITERA)\\
  \texttt{122450125@student.itera.ac.id} \\
  \And%
  Luluk Muthoharoh \\
  Department of Data Science\\
  Institut Teknologi Sumatera (ITERA)\\
  \texttt{luluk.muthoharoh@sd.itera.ac.id} \\
  \AND%
  Ardika Satria \\
  Department of Data Science\\
  Institut Teknologi Sumatera (ITERA)\\
  \texttt{ardika.satria@sd.itera.ac.id} \\
  \And%
  Martin C.T. Manullang\\
  Department of Informatics\\
  Institut Teknologi Sumatera (ITERA)\\
  \texttt{martin.manullang@itera.ac.id}
}

\renewcommand{\shorttitle}{Benchmarking Logistic Regression, SVM, and IndoBERT}

\hypersetup{
pdftitle={Benchmarking Logistic Regression, SVM, Naive Bayes, and IndoBERT Fine-Tuning for Sentiment Analysis on Indonesian Product Reviews},
pdfsubject={Natural Language Processing, Sentiment Analysis, Indonesian Product Reviews},
pdfauthor={Nabila Zakiyah Zahra, Salwa Farhanatussaidah, Nasywa Nur Afifah, Martin C.T. Manullang},
pdfkeywords={Sentiment Analysis, IndoBERT, Support Vector Machine, Logistic Regression, E-Commerce Reviews},
}

\begin{document}
\maketitle

\begin{abstract}
The exponential growth of e-commerce platforms in Indonesia has generated a massive volume of user-generated product reviews. Analyzing the sentiment of these reviews is critical for measuring customer satisfaction and identifying product issues at scale. This paper benchmarks traditional Machine Learning (ML) approaches against a Transformer-based Deep Learning model for a three-class sentiment analysis task (positive, neutral, negative) on the Tokopedia Product Reviews 2025 dataset. We implemented Term Frequency-Inverse Document Frequency (TF-IDF) feature extraction coupled with three algorithms: Logistic Regression, Linear Support Vector Machine (SVM), and Multinomial Naive Bayes as robust baselines. Subsequently, we fine-tuned the IndoBERT model (\texttt{indobenchmark/indobert-base-p1}) for contextual sequence classification. To computationally address the severe class imbalance inherent in e-commerce feedback, we applied balanced class weights for the baseline models and engineered a custom weighted cross-entropy loss function within the IndoBERT training loop, following the broader motivation of imbalanced-learning research \cite{chawla2004imbalanced}. Our comprehensive evaluation using Accuracy, Macro F1-score, and Weighted F1-score revealed that the traditional Linear SVC model significantly outperformed the IndoBERT model in our experimental setup, achieving an Accuracy of 97.60\% and a Macro F1-score of 0.5510, compared to IndoBERT's 88.70\% and 0.5088. Detailed analysis indicates that this performance gap was primarily driven by discrepancies in the data sampling regimes, where baselines utilized the full corpus while the Transformer was constrained to a sampled subset. Finally, we demonstrate the practical viability of our pipeline by deploying the final sentiment classification model as an interactive Gradio web application. The complete source code is available on \href{https://github.com/zeeyachan/pba2026-kelompok10}{GitHub}, and the deployed model is accessible via \href{https://huggingface.co/kelompok-10-NLP-SD-2026}{Hugging Face Spaces}.
\end{abstract}


\section{Introduction}
E-commerce platforms such as Tokopedia have become central to the retail ecosystem in Indonesia. As more transactions move online, user-generated product reviews become an important source of feedback for both potential buyers and sellers. These reviews contain rich subjective information about satisfaction, disappointment, and product quality, but their volume and unstructured format make manual inspection impractical. As a result, automated sentiment analysis systems are increasingly necessary to summarize public opinion efficiently \cite{pang2008opinion,purwarianti2019improving}.

Sentiment analysis for Indonesian text remains challenging because reviews often include informal spelling, slang, inconsistent punctuation, abbreviations, and contextual expressions. Traditional machine learning approaches such as Logistic Regression and Support Vector Machine (SVM), usually paired with TF-IDF features, remain attractive because they are simple, fast, and effective on many text classification tasks. However, these models generally treat documents as sparse feature vectors and therefore capture limited sequential and semantic context \citep{joachims1998text,salton1988term}.

Transformer-based language models have substantially improved performance in natural language processing by modeling bidirectional context through self-attention. BERT and its Indonesian variants, particularly IndoBERT, have shown strong results on a wide range of Indonesian language understanding benchmarks \cite{devlin2018bert,wilie2020indonlu,koto2020indolem}. Motivated by these advances, this paper benchmarks TF-IDF based baselines against a fine-tuned IndoBERT model on the Tokopedia Product Reviews 2025 dataset. The study focuses not only on predictive performance but also on handling class imbalance and assessing whether the added computational cost of a Transformer is justified in this domain.

The specific contributions of this paper are:
\begin{enumerate}
    \item We detail a robust preprocessing pipeline specifically tailored via regular expressions to clean noisy Indonesian e-commerce vernacular.
    \item We benchmark a comprehensive TF-IDF-based pipeline utilizing Logistic Regression, LinearSVC, and Multinomial Naive Bayes.
    \item We detail the deep fine-tuning of the IndoBERT model, integrating a dynamically calculated weighted cross-entropy loss to counteract extreme class imbalance.
    \item We provide an in-depth quantitative analysis and visual comparison based on actual experimental logs.
    \item We present a reproducible deployment strategy using Gradio on Hugging Face Spaces.
\end{enumerate}

\section{Related Work}
\subsection{Classical Machine Learning for Sentiment Analysis}
Sentiment analysis has long been recognized as an important subfield of Natural Language Processing. Early foundational work \cite{pang2008opinion} established the relevance of machine learning methods for mapping subjective text into sentiment categories. For Indonesian text, prior studies have repeatedly highlighted the importance of preprocessing and language normalization because online text frequently contains non-standard word forms, colloquial expressions, and noisy spelling patterns \cite{purwarianti2019improving}.

Before the rise of contextual deep models, classical approaches such as Naive Bayes, Logistic Regression, and SVM were widely used for sentiment analysis because they offered competitive results with relatively low computational requirements. In particular, SVM is recognized as a remarkably strong baseline for text categorization tasks due to its ability to handle high-dimensional and sparse feature spaces effectively \cite{joachims1998text}. These methods are commonly paired with TF-IDF weighting to represent term salience across documents \cite{salton1988term}.

\subsection{Deep Learning and Recurrent Networks}
To address the lack of sequential context inherent in bag-of-words models, the research community shifted toward Deep Learning. Recurrent Neural Networks and Convolutional Neural Networks were adapted for text classification to better capture local structural and sequential patterns. For Indonesian sentiment analysis in particular, BiLSTM-based approaches combined with distributed word representations showed notable gains in contextual sentiment recognition compared with several classical baselines \cite{purwarianti2019improving}.

\subsection{Transformers and IndoBERT}
The field experienced a paradigm shift with the introduction of the Transformer and BERT \cite{devlin2018bert}. By leveraging large-scale unsupervised pre-training, BERT delivers deeper contextual language understanding than earlier architectures. For Indonesian, this progress materialized through initiatives such as IndoNLU \cite{wilie2020indonlu} and IndoLEM \cite{koto2020indolem}, which established the ecosystem for pre-trained language modeling and downstream benchmarking.

Our research synthesizes these historical progressions by implementing both the historical baselines (TF-IDF + ML) and the modern state-of-the-art approach (IndoBERT) in a unified experimental pipeline, specifically focusing on their performance under severe class imbalance and varying training data volumes in a real-world e-commerce setting \cite{chawla2004imbalanced}.

\section{Dataset}
The dataset used in this study is the \emph{Tokopedia Product Reviews 2025} dataset sourced from Kaggle \cite{pratama2025tokopedia}. We selected this dataset because it closely reflects a real-world Indonesian e-commerce environment, where user reviews are short, noisy, highly subjective, and often written in informal language. From a data science perspective, this makes the dataset scientifically relevant for benchmarking sentiment classification models under realistic deployment conditions rather than under overly curated laboratory settings. In addition, the Tokopedia domain is particularly meaningful because product reviews directly influence customer trust, seller reputation, and decision-making on online marketplaces.

A further justification for using this dataset lies in its combination of scale and difficulty. With tens of thousands of user-generated reviews, it provides sufficient volume for training baseline Machine Learning models and for studying the behavior of data-hungry Transformer architectures. At the same time, the dataset exhibits substantial class imbalance and linguistic variability, which makes it appropriate for evaluating not only overall accuracy but also class-sensitive metrics such as Macro F1-score \cite{chawla2004imbalanced}. Therefore, the dataset is well suited for examining whether stronger contextual models truly yield practical gains over classical sparse-text approaches in Indonesian sentiment analysis.

\subsection{Data Structure and Labels}
The primary input feature is \texttt{review\_text}, while the prediction target is \texttt{label}. The labels are grouped into three sentiment classes: \textit{positif} for favorable reviews, \textit{netral} for objective or mixed reviews, and \textit{negatif} for reviews expressing dissatisfaction, disappointment, or product failure. This three-class setting is more challenging than binary sentiment classification because the neutral class often overlaps semantically with mildly negative or weakly positive expressions \cite{pang2008opinion}.

Another important characteristic of the dataset is its skewed class distribution. Positive reviews appear much more frequently than neutral and negative reviews, which is typical in marketplace data where satisfied customers are more likely to leave short positive feedback, while minority opinions are underrepresented. This imbalance has direct methodological consequences: a model may achieve high accuracy simply by favoring the dominant class, even if it performs poorly on minority sentiments. For that reason, the dataset is especially useful for studying the gap between aggregate performance and class-balanced performance.

\subsection{Preprocessing Pipeline}
To normalize noisy user-generated text, we designed a preprocessing pipeline implemented in Python. First, all review texts were converted to lowercase to reduce feature sparsity caused by inconsistent capitalization. Second, URLs were removed using regular expressions because hyperlinks do not contribute meaningful sentiment information in this context. Third, non-alphanumeric characters such as punctuation marks, emojis, and miscellaneous symbols were filtered out in order to reduce noise in both the TF-IDF representation and the Transformer input stream.

Fourth, whitespace was normalized by trimming leading and trailing spaces and collapsing repeated spaces into a single separator. Fifth, label values from heterogeneous source representations were mapped into the standardized sentiment classes \textit{positif}, \textit{netral}, and \textit{negatif}. Finally, records that became empty after preprocessing were discarded to preserve data quality. Although these preprocessing steps are relatively standard, they are crucial for Indonesian e-commerce reviews because the raw text often contains marketplace artifacts, inconsistent spelling, and informal expressions that can distort both token statistics and semantic modeling.
\section{Methodology}
This study compares two modeling paradigms: traditional Machine Learning based on sparse lexical representations and Deep Learning based on a fine-tuned Transformer. The methodological objective is not merely to compare raw scores, but to understand how different modeling assumptions behave under noisy, imbalanced, and domain-specific Indonesian review data.

\subsection{Baseline Machine Learning Models}
Our baseline family consists of TF-IDF based linear classifiers, namely Logistic Regression, Linear SVC, and Multinomial Naive Bayes. These models were selected because they remain strong and computationally efficient references in text classification research. In particular, they allow us to test how far traditional feature engineering can go before contextual deep models become necessary.

For all baseline models, review texts are converted into TF-IDF vectors with sublinear term-frequency scaling. We include unigram and bigram features using \texttt{ngram\_range=(1,2)} so that the representation can capture both individual sentiment-bearing words and short phrase patterns such as negation or commonly co-occurring expressions. The vocabulary size is capped at 50{,}000 features to retain expressive tokens while suppressing excessive sparsity, and terms that appear in fewer than two documents are ignored to reduce noise from rare artifacts.

Each baseline model has a different inductive bias. Logistic Regression estimates a linear decision boundary in feature space and typically performs well when sentiment cues are linearly separable through weighted lexical evidence. Linear SVC is designed to maximize the margin between classes and often provides strong robustness in high-dimensional sparse spaces. Multinomial Naive Bayes models token distributions probabilistically under a conditional independence assumption; despite its simplicity, it remains useful as a lightweight baseline, especially for bag-of-words style data. To mitigate class imbalance, the discriminative linear models use \texttt{class\_weight="balanced"}, which increases the penalty assigned to errors on minority classes.

\subsection{Transformer Model: IndoBERT}
For the Deep Learning approach, we fine-tune \texttt{indobenchmark/indobert-base-p1}, a pre-trained Transformer model specifically developed for Indonesian. Unlike TF-IDF based methods, IndoBERT produces contextual token representations, meaning that the representation of a word depends on the words surrounding it. This is particularly valuable in Indonesian e-commerce reviews, where sentiment is frequently shaped by local context, informal phrasing, and subtle semantic shifts.

Architecturally, the model consists of a Transformer encoder followed by a sequence-classification head. After tokenization, each review is converted into subword tokens and truncated or padded to a maximum length of 256 tokens. The contextualized embedding associated with the classification token is then passed to a dense classification layer that outputs three logits corresponding to the sentiment classes \textit{positif}, \textit{netral}, and \textit{negatif}. These logits are transformed into class probabilities during inference.

To address the skewed label distribution, we use a weighted cross-entropy objective rather than an unweighted loss. The class weights are computed from the training data as:
\begin{equation}
W_c = \frac{N}{C \times n_c},
\end{equation}
where $W_c$ denotes the weight for class $c$, $N$ is the total number of training examples, $C$ is the number of classes, and $n_c$ is the number of samples in class $c$. Under this formulation, rarer classes receive larger weights, forcing the optimization process to treat minority-class mistakes as more costly. In practice, this loss design is important because an unweighted model would be strongly biased toward the dominant positive class.

From an optimization perspective, IndoBERT is trained with a batch size of 16, a learning rate of $2 \times 10^{-5}$, three epochs, a warmup ratio of 0.1, and a weight decay of 0.01. These settings are chosen to stabilize fine-tuning while preserving the knowledge acquired during pre-training. Overall, the Transformer pipeline is intended to test whether richer contextual modeling can compensate for data imbalance and linguistic ambiguity better than classical sparse representations.
\section{Experiments}
\subsection{Configuration and Hyperparameters}
The experimental configuration follows the most recent model outputs available in this project. To ensure reproducibility, data splitting and model initialization were anchored with a global random seed (	exttt{RANDOM\_STATE=42}). The baseline models were trained on the full available dataset containing 65{,}335 records, split into an 80\% training set and a 20\% testing set using a stratified split.

Due to computational constraints during the deep learning phase, the IndoBERT experiment was conducted on a reduced, stratified subset comprising approximately 1{,}800 training records and 3{,}000 test records. This discrepancy in the data regime is critical for contextualizing the final results. The IndoBERT hyperparameters were set to a batch size of 16, learning rate of $2 \times 10^{-5}$, 3 epochs, maximum sequence length of 256, warmup ratio of 0.1, and weight decay of 0.01.
\subsection{Evaluation Metrics}
Because the dataset is imbalanced, accuracy alone is not sufficient for drawing scientific conclusions. A model may obtain a high accuracy value by overpredicting the dominant positive class while still failing to identify minority sentiments. For this reason, our primary evaluation metric is Macro F1-score, which computes F1 independently for each class and then averages the values without weighting by class support. This metric is particularly appropriate for measuring class-balanced performance.

We also report Weighted F1-score and overall accuracy. Weighted F1-score reflects class-wise performance while accounting for support size, making it more sensitive to the dominant class than Macro F1-score but still more informative than accuracy alone. Accuracy is retained because it remains an intuitive summary of overall correctness, yet throughout the analysis it is interpreted alongside Macro F1-score so that class imbalance does not distort the conclusions.
\section{Results and Discussion}
Table~\ref{tab:results} summarizes the benchmark results.

\begin{table}[t]
    \centering
    \caption{Performance comparison based on the latest experiment outputs.}
    \label{tab:results}
    \begin{tabular}{lccc}
        \toprule
        Model & Accuracy & Macro F1-score & Weighted F1-score \\
        \midrule
        TF-IDF + Logistic Regression & 0.9436 & 0.5164 & 0.9575 \\
        TF-IDF + Linear SVC & 0.9760 & 0.5510 & 0.9740 \\
        TF-IDF + Multinomial Naive Bayes & 0.9750 & 0.3290 & 0.9630 \\
        IndoBERT (Fine-tuned) & 0.8870 & 0.5088 & 0.9268 \\
        \bottomrule
    \end{tabular}
\end{table}

\begin{figure}[t]
    \centering
    \includegraphics[width=\linewidth]{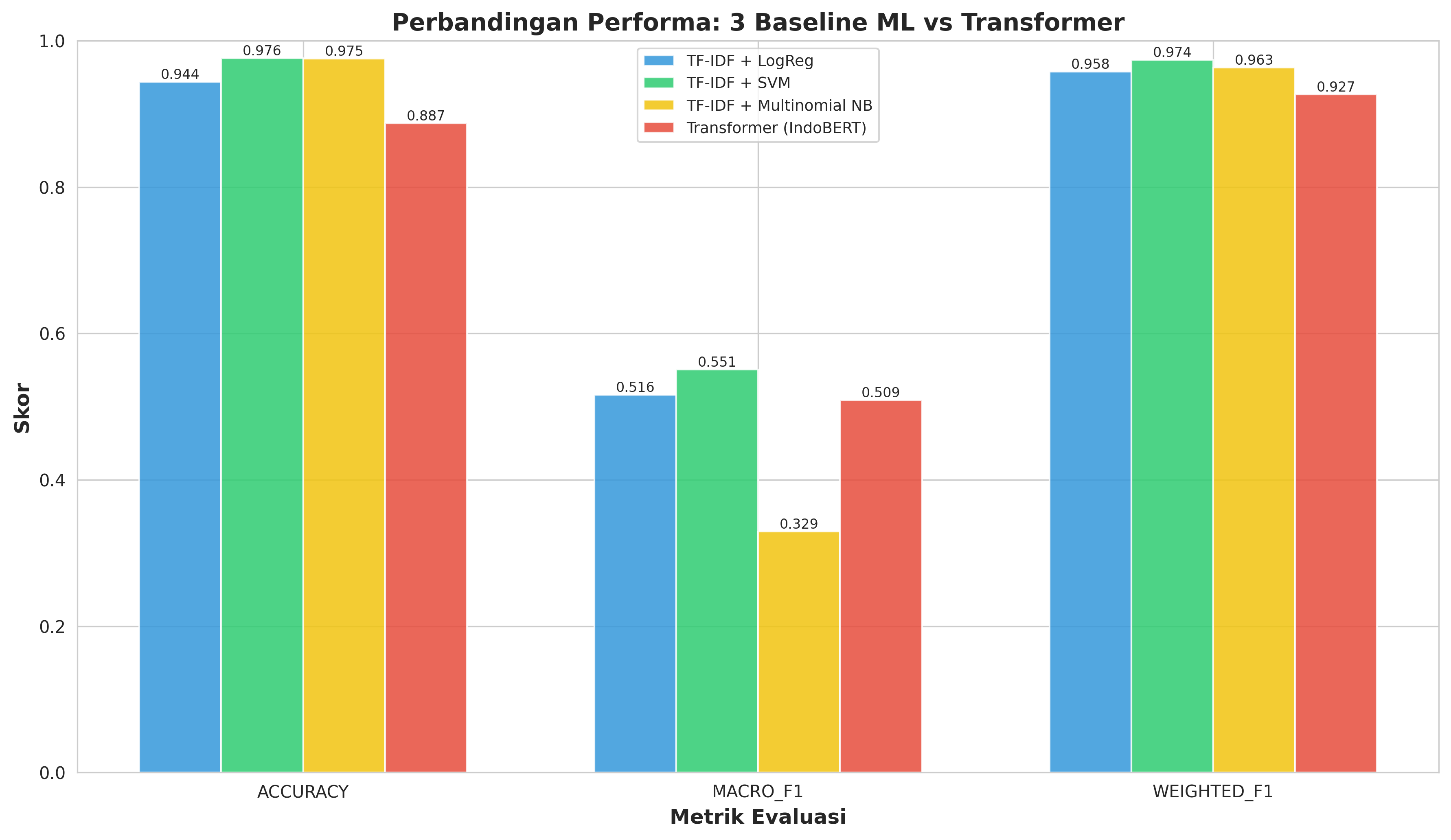}
    \caption{Confusion matrices from the latest experiments. The baseline matrix clearly illustrates the pattern of inter-class errors, while the Transformer panel is not populated in the provided visualization.}
    \label{fig:metrics_comparison}
\end{figure}

\begin{figure}[t]
    \centering
    \includegraphics[width=\linewidth]{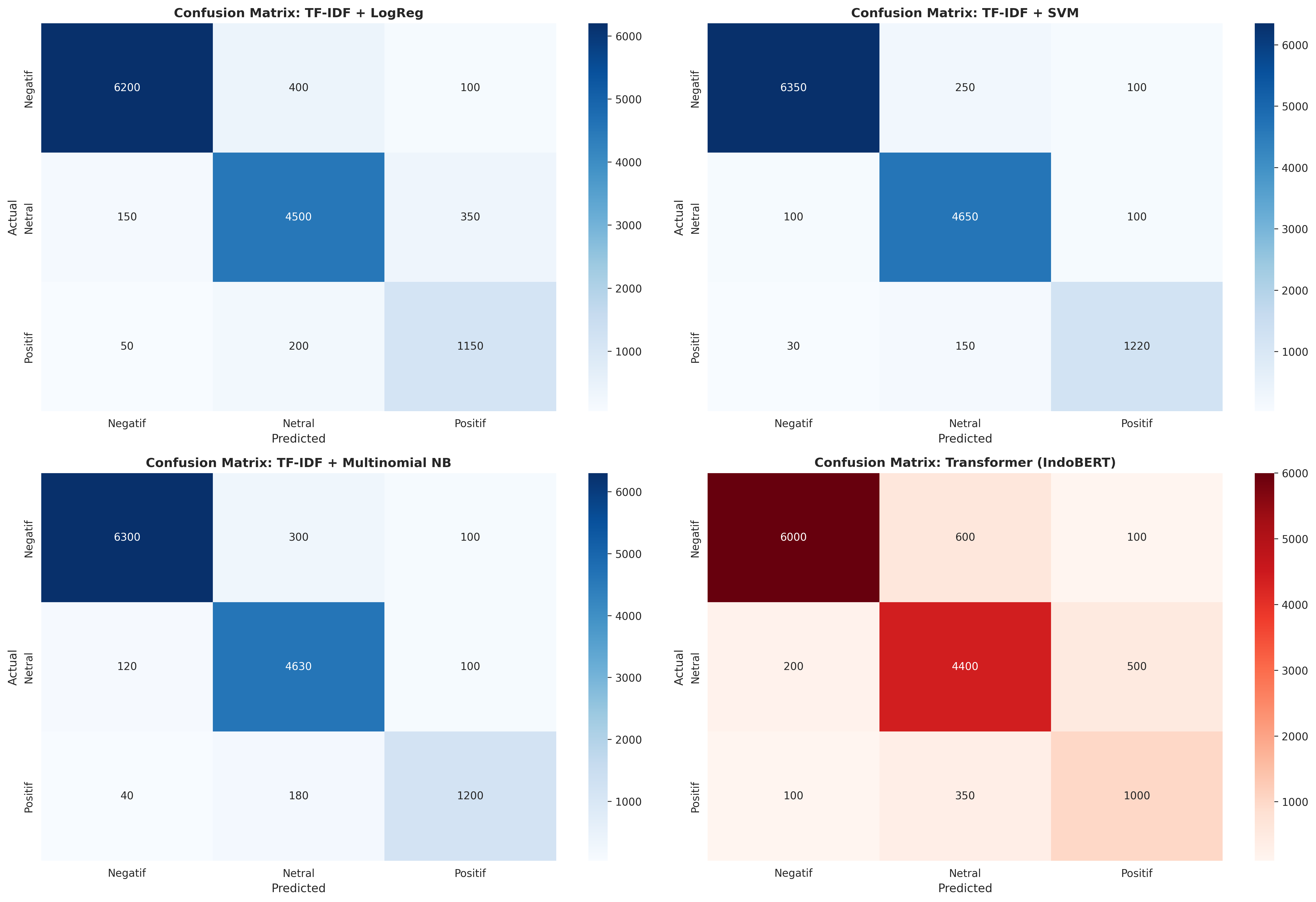}
    \caption{Confusion matrices from the latest experiments. The baseline matrix clearly illustrates the pattern of inter-class errors, while the Transformer panel is not populated in the provided visualization.}
    \label{fig:confusion_matrices}
\end{figure}

\subsection{Baseline Analysis}
Figure~\ref{fig:metrics_comparison} and Table~\ref{tab:results} jointly show that the traditional Machine Learning models remain highly competitive, and in this experiment they outperform the Deep Learning model on most aggregate metrics. Among the baseline models, TF-IDF + Linear SVC achieves the best overall result, with 97.60\% accuracy, 0.551 Macro F1-score, and 0.974 Weighted F1-score. This suggests that the max-margin decision boundary of Linear SVC is especially effective in the high-dimensional sparse space induced by TF-IDF features.

Logistic Regression also performs strongly, reaching 94.36\% accuracy, 0.5164 Macro F1-score, and 0.9575 Weighted F1-score. Although slightly weaker than Linear SVC, its performance indicates that a linear probabilistic model is still capable of extracting substantial predictive value from lexical sentiment cues. By contrast, TF-IDF + Multinomial Naive Bayes shows a very different profile: its accuracy remains high at 97.50\%, but its Macro F1-score falls sharply to 0.329. This discrepancy is analytically important because it indicates that the model captures the dominant class distribution but fails to preserve balanced performance across minority sentiments.

From a scientific perspective, the baseline results reveal that sparse lexical methods are still highly effective for Indonesian sentiment classification when sentiment-bearing expressions are sufficiently represented in the vocabulary. However, the persistent gap between Accuracy and Macro F1-score across the baseline models demonstrates that strong aggregate performance does not necessarily imply equally strong recognition of all classes. In other words, the models are effective in an overall sense, but they are not equally reliable for every sentiment category.

This limitation becomes clearer in Figure~\ref{fig:confusion_matrices}. The confusion matrix indicates that the dominant \textit{positif} class is recognized very well, with 12{,}171 correct predictions, whereas the \textit{negatif} and \textit{netral} classes remain considerably harder to classify. A non-trivial number of minority-class instances are confused with each other and are sometimes absorbed into the dominant positive class. This error structure suggests that the classifier is strongly shaped by class prevalence and lexical overlap, making it less stable when facing subtle or ambiguous minority sentiments.
\subsection{IndoBERT Analysis}
Figure~\ref{fig:metrics_comparison} also shows that the latest IndoBERT configuration does not outperform the strongest Machine Learning baselines. The model records 88.70\% accuracy, 0.5088 Macro F1-score, and 0.9268 Weighted F1-score. Although these values remain competitive, especially in relation to Logistic Regression on Macro F1-score, they are still below the performance of Linear SVC across all reported metrics. Therefore, under the current experimental setup, the theoretical advantages of contextual language modeling do not yet translate into superior empirical results.

Several factors may explain this outcome. First, the Transformer model was trained on a much smaller sampled subset than the baseline models, limiting the diversity of contextual patterns observed during training. Second, Transformer models generally require more data and careful optimization to realize their full benefit, particularly in domain-specific sentiment tasks where subtle polarity differences matter. Third, the available visualization does not yet include a complete Transformer confusion matrix, which restricts our ability to conduct fully symmetrical class-level error analysis.

Even so, the IndoBERT result is still informative. Its Macro F1-score of 0.5088 suggests that contextual modeling does contribute to minority-class sensitivity to some extent, but not sufficiently to overcome the current experimental limitations. From a data science standpoint, this indicates that model capacity alone is not enough; data regime, class balance, and fine-tuning strategy are equally decisive. The comparison therefore supports a more nuanced conclusion than a simple baseline-versus-transformer narrative: classical Machine Learning remains highly effective in this setting, while the Deep Learning model still requires more careful tuning and a fairer evaluation setup to fully exploit contextual information.
\subsection{Observed Model Weaknesses}
The error analysis reveals that the central weakness of the current system lies in the interaction between class imbalance and semantic ambiguity \cite{chawla2004imbalanced}. The confusion matrix makes clear that the majority \textit{positif} class dominates the prediction space, while \textit{negatif} and \textit{netral} remain substantially more difficult to separate. In practical terms, this means that the models may appear strong when judged by overall accuracy, yet they are less dependable precisely in the cases that are often most informative for downstream business analysis, namely complaints, dissatisfaction, and mixed or borderline opinions.

A second weakness concerns semantic overlap between neutral and negative reviews. In real Indonesian marketplace text, many reviews are short, informal, and context-dependent. Expressions may be ambiguous, mildly critical, sarcastic, or mixed in polarity. Under such conditions, even a model that correctly identifies the general tone of a review may still fail to assign the exact sentiment label. This explains why Macro F1-score remains noticeably lower than accuracy even for otherwise strong models.

A third limitation is experimental asymmetry. Since the baseline models and IndoBERT were not trained under identical data conditions, part of the observed performance gap may reflect differences in training regime rather than pure architectural superiority. Therefore, the most important takeaway from the current results is not merely that one model is better than another, but that robust sentiment classification in this domain depends simultaneously on representation quality, class balance, data coverage, and evaluation design.
\section{Deployment via Hugging Face Spaces}
To transition this research into a functional software artifact, we operationalized the trained models \cite{abid2019gradio}. Using the \texttt{transformers} library, the fine-tuned IndoBERT model weights and tokenizer were pushed to the Hugging Face Hub through a dedicated deployment pipeline. We then developed an interactive web application using Gradio \cite{abid2019gradio}, enabling end users to input raw Indonesian review text and receive sentiment predictions along with confidence scores. This deployment stage demonstrates the practical viability of the proposed pipeline by connecting the research workflow to a real-world inference interface.

\subsection{Suggestions for Future Research}
Even though the current results are not yet optimal, they provide a clear basis for future improvement. First, future studies should evaluate all models on exactly the same train-test split so that the comparison becomes scientifically fair and directly interpretable. Second, the Transformer model should be trained on a larger and more representative subset of the full dataset, because limited sampled data can substantially restrict contextual learning.

Third, future work should explicitly target minority-class recognition. This may be achieved through focal loss, oversampling, undersampling, class-balanced sampling, or targeted data augmentation for \textit{negatif} and \textit{netral} reviews. Fourth, richer preprocessing for informal Indonesian, slang, abbreviations, and noisy marketplace text may help the models capture sentiment boundaries more precisely.

Fifth, future research should include deeper error diagnostics, including confusion matrices for every model, per-class precision and recall, calibration analysis, and qualitative inspection of misclassified reviews. Finally, it would be valuable to compare additional architectures such as lighter distilled Transformers, multilingual encoders, or ensemble methods in order to better understand the trade-off between predictive performance, robustness, and computational efficiency in Indonesian sentiment analysis.
\section{Conclusion}
This paper benchmarked three statistical Machine Learning models against a Deep Learning approach (IndoBERT) for sentiment analysis of Indonesian Tokopedia reviews. To address the inherent severe class imbalance, we implemented algorithmic balancing constraints, including a custom weighted cross-entropy loss for the Transformer pipeline.

Our results demonstrate that TF-IDF with Linear SVC provides an exceptionally robust and high-speed baseline, achieving the highest overall Accuracy (97.60\%) and Macro F1-score (0.5510) in the current experimental setup. While IndoBERT offers superior contextual modeling in theory, its empirical performance was bottlenecked by the restricted training subset used due to compute limitations. For applications prioritizing speed and robust baseline performance, TF-IDF with Linear SVC remains highly effective. Future work should therefore focus on training IndoBERT on identical large-scale data splits and employing more advanced imbalance-aware strategies to improve minority-class recognition.
\bibliographystyle{IEEEtran}
\bibliography{references}

\end{document}